# MS-BACO: A NEW MODEL SELECTION ALGORITHM USING BINARY ANT COLONY OPTIMIZATION FOR NEURAL COMPLEXITY AND ERROR REDUCTION

**SAMAN SADEGHYAN**

*Department of Computer Engineering, University of Tehran*
*Tehran, 1417466191, Iran*

**SHAHROKH ASADI**

*Department of Computer Engineering, University of Tehran*
*Tehran, 1417466191, Iran* [*]

Stabilizing the complexity of Feedforward Neural Networks (FNNs) for the given approximation task can be managed by defining an appropriate model magnitude which is also greatly correlated with the generalization quality and computational efficiency. However, deciding on the right level of model complexity can be highly challenging in FNN applications. In this paper, a new Model Selection algorithm using Binary Ant Colony Optimization (MS-BACO) is proposed in order to achieve the optimal FNN model in terms of neural complexity and cross-entropy error. MS-BACO is a meta-heuristic algorithm that treats the problem as a combinatorial optimization problem. By quantifying both the amount of correlation exists among hidden neurons and the sensitivity of the FNN's output to the hidden neurons using a sample-based sensitivity analysis method called, extended Fourier amplitude sensitivity test, the algorithm mostly tends to select the FNN model containing hidden neurons with most distinct hyperplanes and high contribution percentage. Performance of the proposed algorithm with three different designs of heuristic information is investigated. Comparison of the findings verifies that the newly introduced algorithm is able to provide more compact and accurate FNN model.

## 1. Introduction and literature review

The most compelling approach is to think of Feed-forward Neural Networks (FNNs), as function approximation machines. They are designed to learn complex nonlinear function mapping of some set of input values to output values. In the context of deep learning, FNNs are the quintessential models that have been broadly used as a powerful machine learning algorithm in various fields of science and engineering, e.g. identification and control of dynamical systems [1], robotics [2], forecasting financial and economic time series [3], renewable power systems [4], big data [5].

---
[*]Corresponding author
  Email addresses: s.asadi520@gmail.com (S. Asadi),
sadeghyan95@ut.ac.ir (S. Sadeghyan).

Every hidden layer provides a new representation of the input data. The more complexity exists in working dataset, the more complex representation of the data (larger hidden layer), is needed such that allows the network to be able to properly learn the desired function [6-8]. As an up-close example regarding this issue, three different classifier FNNs with a single hidden layer composing of respectively 2, 4 and 20 hidden neurons are trained over an artificially generated dataset. The dataset is made of a two-dimensional vector divided into two non-linearly separable class labels. For the purpose of comparison, the hyperplanes of hidden neurons and the produced global decision surface of each network along the test data are plotted in Fig. 1. Ultimately, we come up with multiple opinions:

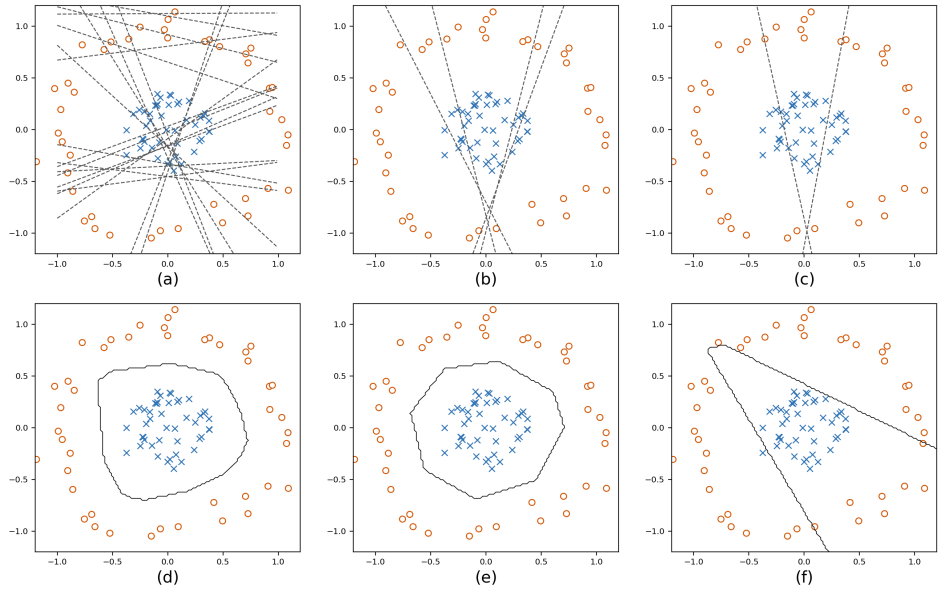

Fig. 1. The comparison of hidden neuron hyperplanes and generalization ability of three different classifier FNNs with a single hidden layer. Each pair of (a, d), (b, e) and (c, f) is the visualized hyperplanes of neurons and the decision surface for the networks with respectively of 2, 4 and 20 hidden neurons that are trained over an artificially generated dataset.

- The complexity of the network mainly influences the generalization quality of the model. Notably, too small network is not able to properly fit the true function described by the training data. In our case study, both larger networks can achieve the classification rate of 100%, while this rate in third network with 2 hidden neurons is 90%, inasmuch as it cannot produce a hypersurface that partitions the underlying vector space into two sets, one for each class.
- Another key point is computational efficiency. The network with 20 hidden neurons which is clearly larger than necessary, requires unneeded arithmetic calculations and in effect more computational resources. But the network with 4 hidden neurons is more efficient in both forward and backward propagations while it is capable of reaching the same classification accuracy. This later becomes more important specially because of the fact that back-propagation as the most popular gradient-based learning algorithm cannot satisfy growing real-time learning needs in many applications [9].

- A network with larger hidden layer has more weight connections which produces more dimensions in weight-space. As a result, more paths are created around the barriers of poor local minima in the lower dimensional subspaces. Thus, local minima problem seemingly is intensified in case of too small networks [10]. On the other hand, Fig. 1.a depicts that many hidden neurons in the large network have very similar or identical hyperplanes. In fact, this similarity might result in redundancy of hidden neurons.

- The complexity increases in an excessively large network, and even though it might be able to accurately approximate the desired function, nevertheless since a larger network learns quicker, it is more likely to have poor generalization due to overfitting [6, 11]. Consequently, learning process would be in further need of using various regularization techniques, which do not always lead to the best solutions.

If the generalization ability of two FNN models trained over the same training data is the same as each other, then the model with simpler structure (lower number of free parameters), should be selected as the best model. Altogether, the network with 4 hidden neurons is identified to have finer level of complexity compared to other models in our experience. But, avoiding overestimating or underestimating size of the network in the applications of FNN is recognized to be a difficult task. Neural network architectures that perform well, are still typically designed manually by experts in a cumbersome trial-and-error process [12], While attempting to find the "minimal" architecture is usually NP-hard [13]. In recent years different algorithms based on techniques such as, sensitivity measure [14-20], correlation [17, 21-25], space-state search [26-30], extreme learning machine [31-34], cascade-correlation [35, 36], sparse signal representation [37] and thermal perceptron rule [38] have been proposed to optimize structure of FNN. A collection of the prominent algorithms is depicted in Fig. 2.

Notably, some researchers have tried to explore the state-space of all different combinations of hidden neurons and layers to find near-optimal solutions. These algorithms make the exploitation more efficient by employing evolutionary computation techniques such as, particle swarm optimization (PSO), and evolutionary algorithm (EA). Apart from evolutionary computation, there are three other fundamental approaches, i.e., node pruning, growing (constructive), and hybrid growing-pruning. The algorithms in Fig. 2 are also tagged with the approach in-use.

In the growing approach, the algorithm starts with a small initial network and gradually adds new hidden units until learning takes place. Although, one difficulty of designing the growing algorithms is defining proper criteria to trigger or stop the procedure of adding new neurons. Otherwise training time escalates and even overfitting may occur [18, 39]. On the other hand, the algorithm using pruning approach starts with a large network and excises the so-called redundant neurons. This approach combines the advantages of training large networks (i.e., learning speed and avoidance of local minima), and small ones (i.e., improved generalization) [40]. Network pruning can be viewed as infinite regularization due to its ability in controlling the model capacity [41]. The hybrid algorithms merge both approaches together, capable of adding or removing neurons at the same time. Albeit, using the growing approach with the mentioned drawbacks alongside network pruning causes undesired fluctuations in estimating the size of the network in the algorithm process that would diminish speed of the algorithm [15]. This paper's approach can be classified as a pruning approach. In the following we briefly address some of related existing works.

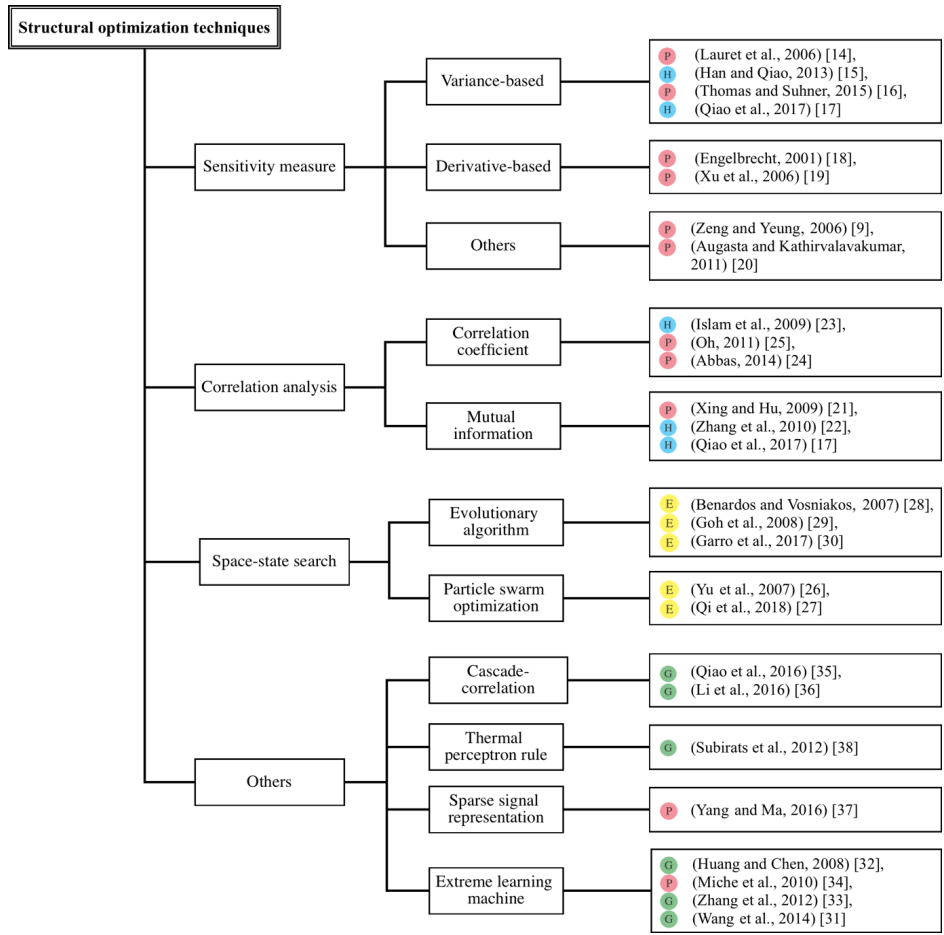

Fig. 2. Taxonomy of the FNN structural optimization techniques, tagged with the approach in-use including, pruning (P), growing (G), hybrid (H) and evolutionary computation (E). The following references are cited in the artwork: [14-38].

A novel pruning algorithm that uses a derivative-based sensitivity analysis technique is presented by Engelbrecht [18], to quantify the relevance of hidden units. Although, the analysis is remained naturally local. Zeng and Yeung [11] also proposed a new technique which prune hidden neurons off the model with the least relevance which is determined by calculating a quantified sensitivity measure. The sensitivity of an individual neuron is defined as the expectation of its output deviation due to expected input deviation with respect to overall inputs from a continuous interval, and the relevance of the neuron is defined as the multiplication of its sensitivity value by the summation of the absolute values of the outgoing weights. Lauret et al. [14] have proposed a new technique based on the global sensitivity analysis in fully connected single layer network. A sample-based method called, Extended Fourier Amplitude Sensitivity Test is utilized to rank hidden units by their contribution percentage to the network output. At the end of each training interval, units which their influence on variation of the network output cannot reach a predefined threshold, are eliminated. Thomas and Suhner [16] also used sensitivity analysis to detect useless units (hidden neurons, inputs, and weights), to prune them afterwards. Han and Qiao [15] and Qiao et al. [17] eliminate neurons based on their contribution derived from the variance-based sensitivity analysis concerning a pre-defined threshold.

As we have seen earlier in the example (Fig. 1), the similarity might result in redundancy of hidden neurons which can be specified by the amount of correlation exists between neurons. Particularly, Islam et al. [23] designed a hybrid algorithm called adaptive merging and growing, which employs the correlation between hidden neurons to merge highly correlated neurons together. Oh [25] and Abbas [24] also remove neurons based on correlation coefficients. Qiao et al. [17] and Zhang et al. [22] used a mutual information criterion to prune hidden neurons by merging correlated hidden neurons. Moreover, Xing and Hu [21] proposed a novel relevance measure to rank and eventually remove the identified redundant neurons from FNN using mutual information.

In addition to the advantages of discussed algorithms, there are some limitations as well. Many of the proposed algorithms contain some problem dependent threshold parameters that are required to be predefined by user, and they directly influence the prosperity of the algorithm. For instance, Han and Qiao [15], Qiao et al. [17] and Lauret et al. [14] prunes hidden units that have a sensitivity measure of less than a constant, which should be carefully tuned. Although using the sensitivity analysis on the network output is a very powerful technique for removing hidden neurons, but algorithms that use this measure alone, and omit the factor of correlation, might result in redundancy of neurons. Moreover, by not considering the consequences of the structural adjustments, there is no guarantee that removing a neuron only based on the fact that it has a low contribution in the network that merely do not satisfy the criterion or it is too similar to another neuron, leads to model error reduction. This would be even irreparably too destructive regarding the network accuracy as well. In this case, the training time can significantly increase due to the false elimination of the units and their relative weight connections with the adapted values in previous learning epochs. Furthermore, these measures alone, are not sufficient since trying to select the best model (set of hidden neurons), based on either network pruning or network growth, is recognized to be a combinatorial optimization problem [42], demanding the employment of a pertinent technique such as evolutionary computation. Although, the proposed evolutionary algorithms on the other hand do not put these measures into practice as well. Prior to presenting the advanced algorithm of the paper for resolving the mentioned drawbacks, we address some of the inspiring insights from recent studies on biological nervous system.

As the biological neurons provided the inspiration for inception of the artificial neuron yielding the famous perceptron model (McCulloch et al., 1943)[43], neural networks process information and operate in a similar manner to biological nervous system. Research on changes of human brain during skill acquisition by Wenger et al. [44], has revealed major volumetric changes of brain in task-relevant areas. Based on the studies, researchers propose an expansion–renormalization model which indicates that learning-related natural neural processes often follow a sequence of expansion, selection, and renormalization [45, 46]. This seems to be an effective method for brain during learning process to first start learning with larger number of neurons than it needs, then gradually test them for their suitability for the role and realizing by which neuron, it can store or carry the information best. Based on which cells function most efficiently and examining different neural architectures, the best model is determined and many cells are eventually pruned away [44, 47].

Inspired by the optimization mechanism of biological nervous system, a new Model Selection algorithm using Binary Ant Colony Optimization (MS-BACO) is proposed in order to achieve the optimal FNN model concerning neural complexity and cross-entropy error. MS-BACO is an meta-heuristic algorithm that tries to select optimal model from an oversized model by taking advantage from the pivotal characteristics of BACO, a swarm intelligence technique that is initially introduced by Hiroyasu et al. [48], based on the traditional Ant Colony Optimization (ACO) [49]. For full reduction of the hidden layer's size, by quantifying both correlation between hidden neurons and contribution of hidden neurons using a sample-based sensitivity analysis method called, extended Fourier amplitude sensitivity test, MS-BACO algorithm mostly tends to select the FNN model containing hidden neurons with the most distinct hyperplanes and high contribution percentage. As the algorithm, itself is assigned to arbitrarily select number of neurons, it is capable of autonomously optimize the network structure instead of using static problem dependent thresholds. The experimental results show that MS-BACO can select more compact FNN models with better generalization quality compared to other algorithms.

Rest of the paper is organized as follows. In Section 2, in addition to the objective FNN model and related notations, two concepts for analyzing hidden neurons are presented. Section 3 proposes our novel algorithm, MS-BACO. Section 4 discusses the implementation and experimental result and it is followed by concluding remarks in Section 5.

## 2.  FNN model and preliminary concepts

By providing enough hidden units for a FNN model with as few as one hidden layer, using any "squashing" activation function (such as the logistic sigmoid activation function), the network is capable of approximating any measurable function to any desired degree of accuracy [50, 51]. Hence, the network considered in this work without loss of generality is a sigmoidal fully connected feedforward model with one hidden layer that is illustrated in Fig. 3. In the following, we discuss FNN related calculations and notations.

Total number of neurons that are incorporated in input and output layers are respectively, $I$ and $K$, which are specified according to the number of features and class labels in given dataset. Also, $N$ as a determinable value by user denotes the number of units in hidden layer. Assume input matrix to be $G = [X_1, X_2, ..., X_L]$, containing $L$ training samples with dimensions of $I \times L$, then $Y_l^{(0)}$ is the output vector of the input layer for the $l_{th}$ sample can be expressed as:

$$Y_l^{(0)} = X_l, \quad (l = 1, ..., L) \qquad (1)$$

since all of the nodes in hidden layer are connected to every node in input layer as well as output layer, $Y_l^{(1)}$ is the output vector of hidden layer that is computed by:

$$Y_l^{(1)} = h(W^T Y_l^{(0)} + B^{(0)}), \quad (l = 1, ..., L) \qquad (2)$$

where $h$ is the sigmoid activation function for each hidden unit with the form of $h(x) = 1/(1 + \exp(-x))$, $W$ is the matrix of weights between input and hidden layers with dimensions of $I \times N$, $B^{(0)}$ is the vector of biases that is connected from the bias node to each node in the hidden layer.

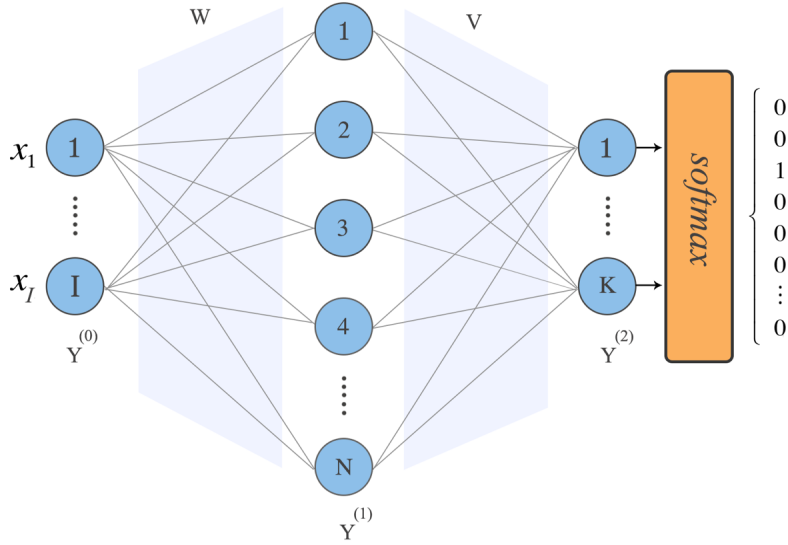

Fig. 3. The sigmoidal fully connected feedforward model with one hidden layer.

In order to network to operate as a classifier model, we want output layer to represent a conditional probability distribution over a discrete variable such as all $K$ target classes in the objective dataset ($C = [c_1, c_2, ..., c_K]$). This is attainable by using the softmax activation function for output layer. Therefore, for $l_{th}$ training sample, while $Y_l^{(2)}$ is output vector of output layer and $P(c_i|X_l)$ represents the probability distribution of the class label $c_i$ (corresponding $i_{th}$ output node), then the calculations of last layer can be expressed as below:

$$Y_l^{(2)} = \text{softmax}(V^T Y_l^{(1)} + B^{(1)}), \quad (l = 1, ..., L) \qquad (3)$$

$$P(c_i|X_l) = \frac{\exp(y_l^{(2)})}{\sum_{k=1}^{K} \exp(y_k^{(2)})}, \quad (k = 1, ..., K) \qquad (4)$$

where $V$ is the matrix of weights between hidden and output layers with dimensions of $N \times K$, $B^{(1)}$ is the vector of biases that is connected from bias node to each node in output layer. $y_i$ and $y_k$ are output values of output unit $i$ and $k$ in $Y_l^{(2)}$. Eventually, using the principle of maximum likelihood determines the predicted class label by the network. For the scheme of training the network in order to learn the output probabilities, we take advantage of Stochastic Gradient Descent (SGD) as one of the most popular gradient-based learning algorithms. It should be noted, for classifier networks with multi-output units, using cross-entropy criterion (i.e., one-hot representation of the label), is proved to be the better approach instead of using mean square error [52, 53]. Altogether, in case of loss function, we try to minimize the Cross-entropy Error (CE) for all training samples:

$$CE = -\frac{1}{L} \sum_{l=1}^{L} \sum_{i=1}^{K} d_i \log P(c_i|X_l) \qquad (5)$$

where $d_i$ is the desired output of the $i_{th}$ output unit, for $l_{th}$ training sample. Meanwhile, using the log function here undoes the exponential of the sigmoid functions in output layer, and prevents the saturation problem in gradient-based learning algorithm [8].

## 2.1. *Sensitivity analysis*

Sensitivity Analysis (SA) is the study of relative importance of different input factors on the model output [54]. Input factors can be any adjustable quantity in specification of the model. A factor can be an initial condition, a parameter, etc. [14]. Performing SA on a complex model can disclose the contribution of each input factor to the variations of the model output. In this work however, the main purpose is to investigate the contribution percentage of each hidden neuron in classification task assigned to the FNN model. Hence, SA provides a saliency measure for hidden neurons. In section 3, we will explain how it is going to be employed in proposed structure optimization algorithm.

Among all of the available methods for SA, sampling method is the most common one that is used in this paper [55]. The procedure of implementing a sample-based method basically starts from specifying the model and input factors that are required to be included in the analysis, and it is followed by defining probability density functions (ranges of variation) for each input factor. After that, the model is evaluated by every sample in a set of samples that are generated accordingly. Eventually, by apportioning the variance of the output according to the input factors, we are able to estimate the contribution of each input factor [56].

For the objective network with $K$ output units, contribution of each hidden neuron in the network is studied with $K$ different models which are illustrated in Fig. 4.

Fig. 4. The specified models included in SA.

Hence, one typical model $k$ with the mapping function $F(.)$ and with respect to $k_{th}$ output unit is specified as follows:

$$Y_k = F(v_{1k} y_1^{(1)}, v_{2k} y_2^{(1)}, \dots, v_{Nk} y_N^{(1)})$$ (6)

In this paper, Extended Fourier Amplitude Sensitivity Test (EFAST), as an efficient sample-based method which has been successfully applied for FNN model [14] is used for generating samples. We also employ Total Effect (TE) as a comprehensive quantitative measure of SA. TE for $n_{th}$ hidden neuron in a typical model like $K$ is computed as bellow:

$$\text{TE}_{nk} = \frac{E_{v_{\sim nk} y_k^{(1)}}[Var(Y|v_{\sim nk} y_k^{(1)})]}{Var(Y)}$$ (7)

where $E_{v_{\sim nk} y_k^{(1)}}[Var(Y|v_{\sim nk} y_k^{(1)})]$ is the expected residual variance of model response when all of the other factors vary but $v_{nk} y_k^{(1)}$, and $Var(Y)$ is the variance of model response (more detailed descriptions regarding TE can be found in [55]).

$S_n^T$ that is sum of computed total effects with respect to all $K$ output units for a neuron is taken into account in order to determine the overall contribution in the network. Eventually, the normalized contribution percentage ($C_n$) for a typical hidden neuron $n$ is expressed as:

$$C_n = \frac{S_n^T}{\sum_{n=1}^{N} S_n^T}$$ (8)

## 2.2. *Correlation analysis*

Correlation is one of the popular statistics that describes the strength of association between two variables. In the hidden layer, we use correlation analysis to reveal the similarity between neurons. The more two neurons are correlated to each other, the more they represent similar functions and hyperplanes in network. Hence, one of them would be recognized to be redundant and might be eliminated. Several techniques have been proposed in statistics to measure correlation. In this study, we use the well-known Pearson product-moment correlation coefficient for quantifying the correlation between hidden neurons. So, correlation coefficient ($R_{ij}$) between two neurons $i$ and $j$ is expressed as:

$$R_{ij} = \frac{\sum_L (y_i^{(1)} - \overline{y}_i)(y_j^{(1)} - \overline{y}_j)}{\sqrt{\sum_L (y_i^{(1)} - \overline{y}_i)^2}\ \sqrt{\sum_L (y_j^{(1)} - \overline{y}_j)^2}} \qquad (9)$$

where, $y_i^{(1)}$ and $y_j^{(1)}$ are output values of hidden neurons $i$ and $j$ for each input sample, and $\overline{y}_i$ and $\overline{y}_j$ denote the mean values of $y_i^{(1)}$ and $y_j^{(1)}$, averaged over L samples. Neurons $i$ and $j$ being perfectly correlated means the exact linear dependency exist and the value of $R_{ij}$ would be 1 or -1. Conversely, $R_{ij}$ being equal to 0 indicates they are completely uncorrelated.

## 3. MS-BACO: a model selection algorithm using binary ant colony optimization

In this section, ideas behind the proposed MS-BACO algorithm are initially outlined. It is followed by describing the framework of the algorithm, given in Algorithm 1. Subsequently, the process and components of the algorithm are discussed in details as well. Obtaining the less complex and more accurate FNN model by searching for the optimal subset of hidden neurons in an extra-large set is the fundamental goal of the proposed algorithm. In this paper, both of the factors discussed in previous section, the sensitivity of FNN's output to the hidden units and the correlation among them help us in investigating the amount of neural redundancy in the objective model.

For the purpose of neuron subset selection, we treat each neuron as a graph node. Thus, the goal becomes finding the shortest path in a graph, where the cost of the paths is cross-entropy error of the model. In other words, the algorithm tries to select the optimal neuron subset in the graph with respect to error reduction.

By using BACO as an evolutionary technique ideal for such combinatorial optimization problem, the search for the low-cost path in the graph is assigned to artificial ants. Ants that mainly emulate their natural counterparts, by utilizing techniques such as pheromone deposit are capable of suggesting good problem solutions. In our case however, the problem solution (selected neuron subset from the set of neurons in hidden layer), is coded as a binary vector with a length equal to the number of hidden neurons. Where for each bit, the value 1 means selecting and 0 means deselecting the corresponding neuron. Particularly, ants in the evolutionary process should assign the best value for binary bits and ultimately suggest the candidate solution. Every node in the graph that ants are traversing is representative of a bit in the solution and it is composed of two sub-nodes, 0 and 1. At each generation, an ant constructs the candidate binary solution by starting from initial node (beginning of the bit string), and traversing all of the nodes in the graph. While visiting a node, an ant determines value of node (bit) by choosing one of sub-nodes as its next move. A typical example of this type of graph with three neurons ($N_1, N_2, N_3$) is illustrated in Fig. 5.

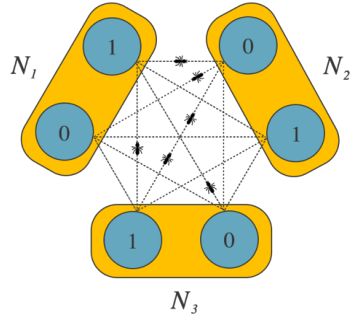

Fig. 5. The graph composed of three neurons ($N_1, N_2, N_3$).

Therefore, every binary solution denotes an FNN model with the corresponding selected hidden neurons. In the following, the process of the proposed algorithm which is given in Algorithm 1 is elaborated in details.

At outset, the input dataset is divided into training, validation and test sets. Then the network is trained using SGD and training samples for e-bet epochs since it is essential to learning to take place prior to SA and correlation analysis [14]. This is followed by computing and setting heuristic values for edges based on that analysis and the specified heuristic design (h-design) (Line 1–6, Algorithm 1). Heuristic information with different designs is discussed in details in sub-section 3.1.

The algorithm continues the process with the evolutionary model selection phase (Line 7–18, Algorithm 1). This phase begins with construction of the binary solutions for m generated ants. The behavior of ants in this phase is described in sub-section 3.2 in details. After pheromone initialization and global update (sub-section 3.3), the best binary solution suggested by ants (best model) is selected for the next iteration.

---

**Algorithm 1.** The general framework of MS-BACO

---

**Input:** dataset, *m*, *N*, *h-design*, *e-bet*, *gen_max*

**Output:** optimum FNN model

1: Split dataset into training, validation and test sets；
2: **while** termination condition is not met **do:**
3:     Train the network with training set for *e-bet* epochs；

4:        Quantify contribution percentage of neurons by Eq. (8)**;**
5:        Compute correlation between all possible pair of neurons by Eq. (9)**;**
6:        Set heuristic values for edges based on input *h-design***;**
7:        Generate m ants $A \leftarrow [\omega_1, \omega_2, ..., \omega_m]$**;**
8:        $\tau_0 \leftarrow 0.1$**;**
9:  **for** $t \leftarrow 1$ to $gen_{max}$ **do:**
10:      **for** each ant $\omega \in A$ **do:**
11:          Randomly place the ant on source node *i* and sub-node *a***;**
12:          **Construction of binary solution $T_\omega$;**
13:      **end for**
14:      Determine $T_{best}$ by evaluating each model using validation set**;**
15:      Pheromone initialization and global update**;**
16:  **end for**
17:  Select the model according to $T_{best}$ for next iteration**;**
18: **end while**
19: Continue the network training until early stopping**;**

At the end of each iteration, it is checked if the termination condition is met or not, which is when the algorithm faces a failure for reducing model size. Eventually the final prompted model is trained until early stopping condition for fine-tuning the network parameters (Line 19).

For better understanding of overall MS-BACO's workflow, all the processing steps are illustrated in Fig. 6.

## 3.1. *Heuristic information (h-design)*

Generally speaking, heuristic value $\eta$ for each edge ending to a sub-node, represents the attractiveness of that sub-node to be traversed by ants. Taking heuristic information into consideration enhances the exploitation ability of the search space that leads to the most promising solutions. Heuristic information can be any subset evaluation function like an entropy based measure or rough set dependency measure [57]. For our objective graph with binary nodes, since each node contains two sub-nodes, there are four edges between each pair of nodes. Fig. 7 illustrates edges and heuristic information between sub-nodes of two typical nodes $N_i$ and $N_j$.

In this paper, before choosing the right design for heuristic information, we assess three different designs (namely, H1, H2 and H3), in order to determine the heuristic values for edges.

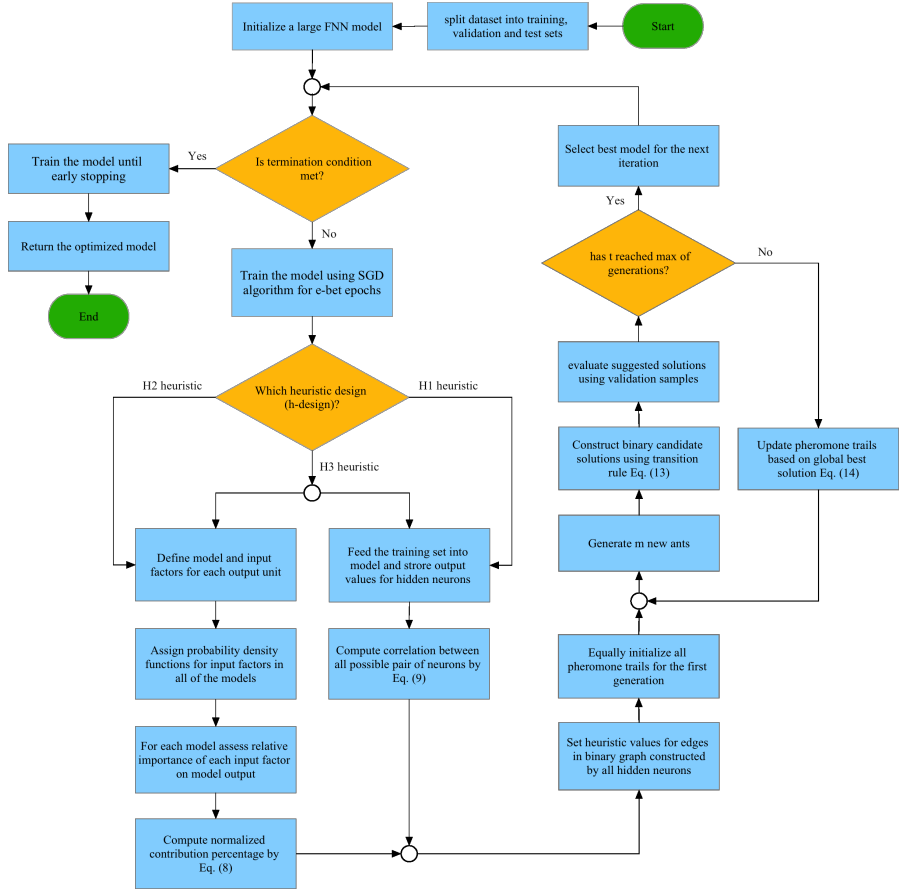

Fig. 6. Workflow of the proposed MS-BACO algorithm.

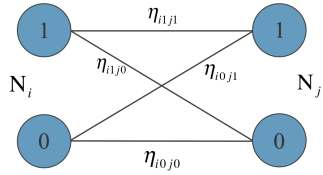

Fig. 7. Edges and heuristic information between sub-nodes.

### *3.1.1 Heuristic design H1*

This design tries to select the most distinct hidden neurons by computing the correlation between them. For instance, assume the correlation between neurons $i$ and $j$ to be high. In this case, hyperplane of two neurons are very similar, and existence of one of them alone would be sufficient in the network. Hence, if one of them is selected, the probability to select or deselect the other neuron can be defined as 1-$|R_{ij}|$ or $|R_{ij}|$. Reversely, if the first neuron is not selected, presence of the other neuron is not necessary too since they have similar functionality. So, the probability to select or deselect the other neuron is defined as 1-$|R_{ij}|$ or $|R_{ij}|$. Eq. (9) describes above statements:

$$\begin{cases} \eta_{i0,j0} = |R_{ij}| \\ \eta_{i0,j1} = 1 - |R_{ij}| \\ \eta_{i1,j0} = |R_{ij}| \\ \eta_{i1,j1} = 1 - |R_{ij}| \end{cases} \qquad (10)$$

### 3.1.2 Heuristic design H2

Significance of hidden neurons in this paper is specified by the contribution percentage ($C_n$) of each neuron, which is computed by using Eq. (8). The aim in this design for computing heuristic values is to urge artificial ants to preferably pick up significant neurons. Assume an ant wants to move from an arbitrary node in the graph by choosing from edges ending to the remaining untraversed nodes. In this case, the more the contribution percentage of a neuron is, the more the probability to select the corresponding node should be, and conversely for less significant neurons. This idea can be expressed as:

$$\begin{cases} \eta_{i0,j0} = (\frac{1}{N}) \sum_{n=1}^{N} C_n \\ \eta_{i0,j1} = C_j \\ \eta_{i1,j0} = (\frac{1}{N}) \sum_{n=1}^{N} C_n \\ \eta_{i1,j1} = C_j \end{cases} \qquad (11)$$

### 3.1.3 Heuristic design H3

In this design for heuristic values, a combination of both previous designs (i.e. significance and similarity of hidden neurons), is considered to determine the desirability of neuron selection. In this case, a neuron that has a high contribution percentage still might not be selected because of a high correlation with another neuron, regardless of its high saliency. Eventually in this combinatorial design, heuristic values on edges are calculated as follows:

$$\begin{cases} \eta_{i0,j0} = |R_{ij}| \ \times \ (\frac{1}{N}) \sum_{n=1}^{N} C_n \\[2mm] \eta_{i0,j1} = (1 - |R_{ij}|) \ \times \ C_j \\[2mm] \eta_{i1,j0} = |R_{ij}| \ \times \ (\frac{1}{N}) \sum_{n=1}^{N} C_n \\[2mm] \eta_{i1,j1} = (1 - |R_{ij}|) \ \times \ C_j \end{cases} \qquad (12)$$

### 3.2. Construction of binary solution $T_\omega$

Artificial ants basically have stochastic behavior similar to any other ACO-based technique. They should make probabilistic decisions based on pheromone trails (i.e., vestiges of previous successful moves), and the problem-specific local heuristic (i.e., reflecting desirability of the move described in 3.1). A typical ant $\omega$ builds the binary solution $T_\omega$, according to this so-called probabilistic transition rule by starting from a random source node and traversing the graph. The procedure of constructing $T_\omega$ that suggests a subset of selected hidden neurons is described in Algorithm 2. Suppose currently ant $\omega$ to be on sub-node $N_{ia}$ ($i = 1, 2, \dots N$ and $a = 0, 1$). The transition probability of the edge connected to a typical sub-node $N_{jb}$ ($j$ is a passable node and $b = 0, 1$), in generation $t$, can be defined as the combination of pheromone density and heuristic desirability of that edge:

$$P_{ia,jb}^{\omega} = \begin{cases} \dfrac{\tau_{ia,jb}^{\alpha} \eta_{ia,jb}^{\beta}}{\sum_n \tau_{ia,n0}^{\alpha} \eta_{ia,n0}^{\beta} \sum_n \tau_{ia,n1}^{\alpha} \eta_{ia,n1}^{\beta}} & \text{if } n \text{ is passable node} \\[4mm] 0 & \text{otherwise} \end{cases} \qquad (13)$$

here $\tau_{ia,jb}$ and $\eta_{ia,jb}$ are respectively the amount of pheromone and heuristic value on edge $(ia, jb)$. $\alpha$ and $\beta$ are two parameters that determine the relative importance of the pheromone value and the heuristic information. In fact, the best balance between exploitation and exploration is achieved by proper values for these parameters.

### 3.3. Pheromone initialization and global update

After all ants completed their tours, suggested solutions are evaluated and the best-so-far tour $T_{best}$ is verified. In the first generation, initial pheromone value $\tau_0$ on all edges is initially set to a same value like 0.1 (Line 8, Algorithm 1). Though before updating pheromone trails at the end of this generation, they are set to $\text{obj}(T_{best})/n$, where $\text{obj}(T_{best})$ is the objective function of $T_{best}$.

For the purpose of updating pheromone trails at the end of each generation (as well as first one), we only allow the global best ant to update the pheromone trails for sake of raising the attractiveness of best solution for next generations. Moreover, pheromone evaporation on all edges is also initiated at this point. This action generally helps to prevent rapid convergence of the algorithm toward a local minimum. Updating pheromone values for the edge $(i, j)$ is performed as follows:

$$\tau_{ij}(t + 1) = (1 - \rho)\tau_{ij}(t) + \rho \times \text{obj}(T_{best}) \qquad (14)$$

where $\rho \in (0, 1]$ is an input parameter called evaporation rate. In this particular method for updating pheromone trails, it can be simply proved that the pheromone trails can never have a value higher than $\text{obj}(T_{best})$, and it is implicitly guaranteed that $\vee (i, j)$: $\tau_0 \leq \tau_{ij} \leq \text{obj}(T_{best})$ [58]. This is preferable, inasmuch as other methods such as max-min ant system is in demand of setting thresholds like maximum and minimum. The objective function in MS-BACO is $\text{obj}(T_w) = 1/\text{CE}$, where CE is the cross-entropy error of the model with given subset of hidden neurons selected in the corresponding $T_w$. This causes the algorithm to choose the model with less classification error.

---

**Algorithm 2.** Construction of binary solution $T_\omega$

---

**Input:** $\omega$, $i$, $a$

**Output:** binary solution $T_\omega$

1: Calculate probability values for all edges by transition rule Eq. (13);
2: **while** there is an unvisited node **do:**
3:     $rand \leftarrow$ generate random value in [0, 1];
4:     $sum \leftarrow 0$
5:     **for** each edge $\in$ [all *2(N-1)* connected edges] **do:**
6:         $sum \leftarrow sum$ + probability value of the edge;
7:         **if** $rand < sum$ **then:**
8:             Move the ant $\omega$ to the selected sub-node;
9:             Set the bit value in $T_\omega$ for corresponding node;
10:             Change probability of all edges attached to this node into 0;
11:         **end if**
12:     **end for**
13: **end while**

---

## 4. Experiments and result Analysis

In this section, we evaluate our proposal by a series of experiments on classification benchmarks. The analysis mainly focuses on specifying the best design for heuristic information and then comparing effectiveness of MS-BACO algorithm with existing related works and FNN with static structure. Proposed algorithm and the three layer FNN model, discussed in section 2 are implemented in IPython system [59], by exploiting open source libraries from scipy stack [60].

### 4.1. *Experimental setup*

Simulations are performed on 14 classification problems belonging to the UCI (University of California, Irvine) machine learning repository [61], with intend to cover examples of small, medium high-dimensional datasets. During each experimentation, 50% of samples were randomly selected for the training set and 25% for validation set and the remaining 25% were for the test set. The exact size of different segments and other properties of these datasets are summarized in Table 1. Validation samples are not fed to the model during the training process, instead they are used for the purpose of early stopping procedure, which prevent model to be overfitted by training samples. The validation set is used for calculating the objective function. Ultimately, the test set is the only set that is used to specify the generalization quality of the model.

Table 1. Properties of datasets used in the experiments from UCI.

| Dataset | Input features | Output classes | Training instances | Validation instances | Test instances | Total |
|---|---|---|---|---|---|---|
| Iris | 4 | 3 | 75 | 37 | 38 | 150 |
| Liver disorders | 6 | 2 | 173 | 86 | 86 | 345 |
| Diabetes | 8 | 2 | 384 | 192 | 192 | 768 |
| Yeast | 8 | 10 | 742 | 371 | 371 | 1484 |
| Breast Cancer | 9 | 2 | 349 | 175 | 175 | 699 |
| Wine | 13 | 3 | 89 | 45 | 44 | 178 |
| Hepatitis | 19 | 2 | 78 | 39 | 38 | 155 |
| Thyroid | 21 | 3 | 3600 | 1800 | 1800 | 7200 |
| Mushroom | 21 | 2 | 4062 | 2031 | 2031 | 8124 |
| Horse colic | 27 | 2 | 184 | 92 | 92 | 368 |
| Ionosphere | 34 | 2 | 176 | 88 | 87 | 351 |
| Arcene | 10,000 | 2 | 450 | 225 | 225 | 900 |

In the following experiments, we normalized the data and rescaled all the features so each feature has a mean of zero and unit variance. Input value for the bias vector is 1 and the network weight connections are initialized with random values in the range [−1, 1] using a random seed. The *e-bet* input parameter which is the given number of training epochs between each model selection phase, as also Lauret et al. [14] mentioned, has not to be carefully tuned for the purpose of hidden neuron analysis such as SA. Based on experiment, the initial number of hidden neurons in input network is 50, and the learning rate in SGD algorithm is set to 0.1.

Furthermore, parameters for ACO are initialized as follows: $\alpha = 1$, $\beta = 0.6$, $\rho = 0.1$. Also, the population size and the maximum number of generations are respectively set to 50 and 30.

### 4.2. *Resolving heuristic information design*

In order to determine the design with the best performance among H1, H2 and H3 and also H0, the state of having no heuristic information ($\beta = 0$), the average testing classification accuracy and the number of remaining hidden neurons in the output model, over 30 independent runs are given in Table 2. The last row of the table shows the average value of given numbers over each experiment. According to this table, MS-BACO with H3 heuristic design could achieve the less complex model with higher accuracy among others. The average CPU time for each experiment is displayed in Table 3. H0 heuristic has the lowest CPU time inasmuch as there is no computational cost for calculating heuristic information. On the other hand, H3 heuristic has the highest CPU time because both sensitivity and correlation analysis should be performed for calculating heuristic values on edges.

Table 2. Performance comparison of MS-BACO algorithm using different heuristic designs. The results are averaged over 30 independent runs. The number below each column of the table shows the average value over all problems.

| Datasets | Classification accuracy (%) | | | | No. of remaining neurons | | | |
|---|---|---|---|---|---|---|---|---|
| | MS-BACO H0 | MS-BACO H1 | MS-BACO H2 | MS-BACO H3 | MS-BACO H0 | MS-BACO H1 | MS-BACO H2 | MS-BACO H3 |
| **Diabetes** | 78.73±1.34 | 79.50±1.10 | 79.54±1.29 | 79.71±1.12 | 5.03±1.74 | 5.23±1.38 | 4.36±1.60 | 4.29±1.42 |
| **Yeast** | 59.20±1.94 | 58.93±1.55 | 59.08±1.56 | 59.14±1.31 | 13.36±1.32 | 12.36±2.10 | 11.26±1.42 | 11.23±1.67 |
| **Breast cancer** | 97.08±0.82 | 97.20±0.79 | 97.51±0.62 | 97.63±0.82 | 2.74±1.05 | 3.13±1.11 | 2.43±0.89 | 2.36±0.79 |
| **Iris** | 98.33±1.00 | 98.31±0.85 | 98.72±0.52 | 98.91±0.40 | 1.96±1.19 | 1.80±0.65 | 2.06±0.85 | 1.86±0.61 |
| **Wine** | 98.51±1.17 | 98.57±1.00 | 98.84±1.13 | 98.94±1.04 | 5.54±1.05 | 5.22±1.23 | 5.51±1.10 | 5.29±1.20 |
| **Hepatitis** | 87.84±2.22 | 87.68±2.41 | 88.32±2.47 | 88.29±2.77 | 5.89±1.38 | 4.60±1.11 | 5.24±1.23 | 4.33±1.53 |
| **Mushroom** | 99.95±0.08 | 99.89±0.02 | 99.97±0.03 | 99.98±0.02 | 5.34±2.84 | 5.33±3.31 | 5.44±3.04 | 5.13±2.98 |
| **Horse colic** | 98.00±1.74 | 98.85±1.65 | 97.94±1.40 | 98.54±1.32 | 7.36±2.54 | 7.60±2.00 | 6.81±2.98 | 7.00±2.55 |
| **Liver disorders** | 76.04±2.75 | 77.19±2.78 | 77.13±3.01 | 77.04±2.74 | 6.49±3.83 | 6.24±3.14 | 5.34±3.02 | 6.11±2.22 |
| **Thyroid** | 97.01±1.50 | 96.73±1.63 | 97.40±1.00 | 97.71±0.85 | 4.88±2.00 | 5.04±2.32 | 5.24±1.40 | 4.93±1.24 |
| **Ionosphere** | 94.88±1.10 | 95.43±1.10 | 95.24±1.10 | 95.24±0.72 | 4.20±1.18 | 3.91±1.75 | 4.21±1.06 | 3.49±1.15 |
| **Arcene** | 89.60±1.64 | 89.23±0.93 | 89.72±1.71 | 90.03±1.44 | 13.85±5.32 | 16.55±4.81 | 15.31±6.34 | 16.75±4.12 |
| **Average** | 89.59 | 89.79 | 89.95 | **90.09** | 6.38 | 6.41 | 6.10 | **6.06** |

The average value of CE in all generations for the first iteration in each experiment is also outlined in Fig. 8-11. As these figures show, H3 heuristic can achieve smaller CE value (i.e., ultimately better objective function), at the final generation in most of the problems. Altogether, for the rest of the experiments H3 design is used for calculating the heuristic information in MS-BACO algorithm.

Table 3. The average execution time (s) of MS-BACO algorithm using different heuristic designs for each problem.

| Datasets | MS-BACO H0 | MS-BACO H1 | MS-BACO H2 | MS-BACO H3 |
|---|---|---|---|---|
| **Diabetes** | 6.93 | 6.74 | 6.72 | 7.04 |
| **Yeast** | 11.35 | 12.16 | 12.04 | 12.40 |
| **Breast cancer** | 4.63 | 4.89 | 5.20 | 5.36 |
| **Iris** | 4.10 | 4.21 | 4.63 | 4.81 |
| **Wine** | 5.60 | 4.22 | 4.62 | 5.08 |
| **Hepatitis** | 5.32 | 6.18 | 4.94 | 5.11 |
| **Mushroom** | 14.57 | 15.00 | 14.55 | 13.36 |
| **Horse colic** | 15.29 | 15.95 | 16.75 | 17.36 |
| **Liver disorders** | 7.24 | 7.23 | 8.20 | 8.24 |
| **Thyroid** | 13.26 | 13.60 | 14.40 | 15.74 |
| **Ionosphere** | 11.94 | 13.58 | 12.10 | 13.54 |
| **Arcene** | 95.00 | 94.25 | 118.30 | 126.34 |
| **Average** | **16.26** | 16.50 | 18.53 | 19.53 |

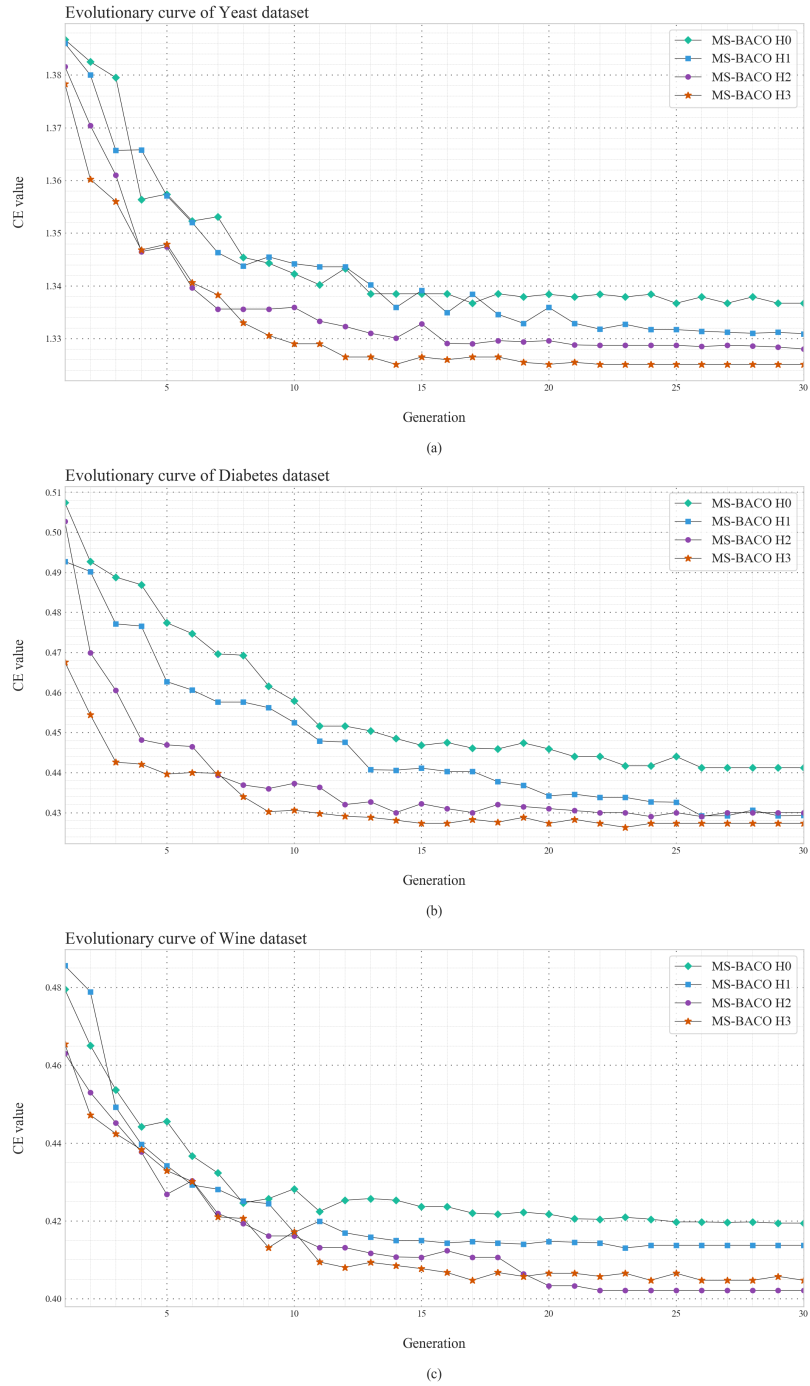

Fig. 8. Average CE of datasets including, Yeast (a), Diabetes (b), Wine (c).

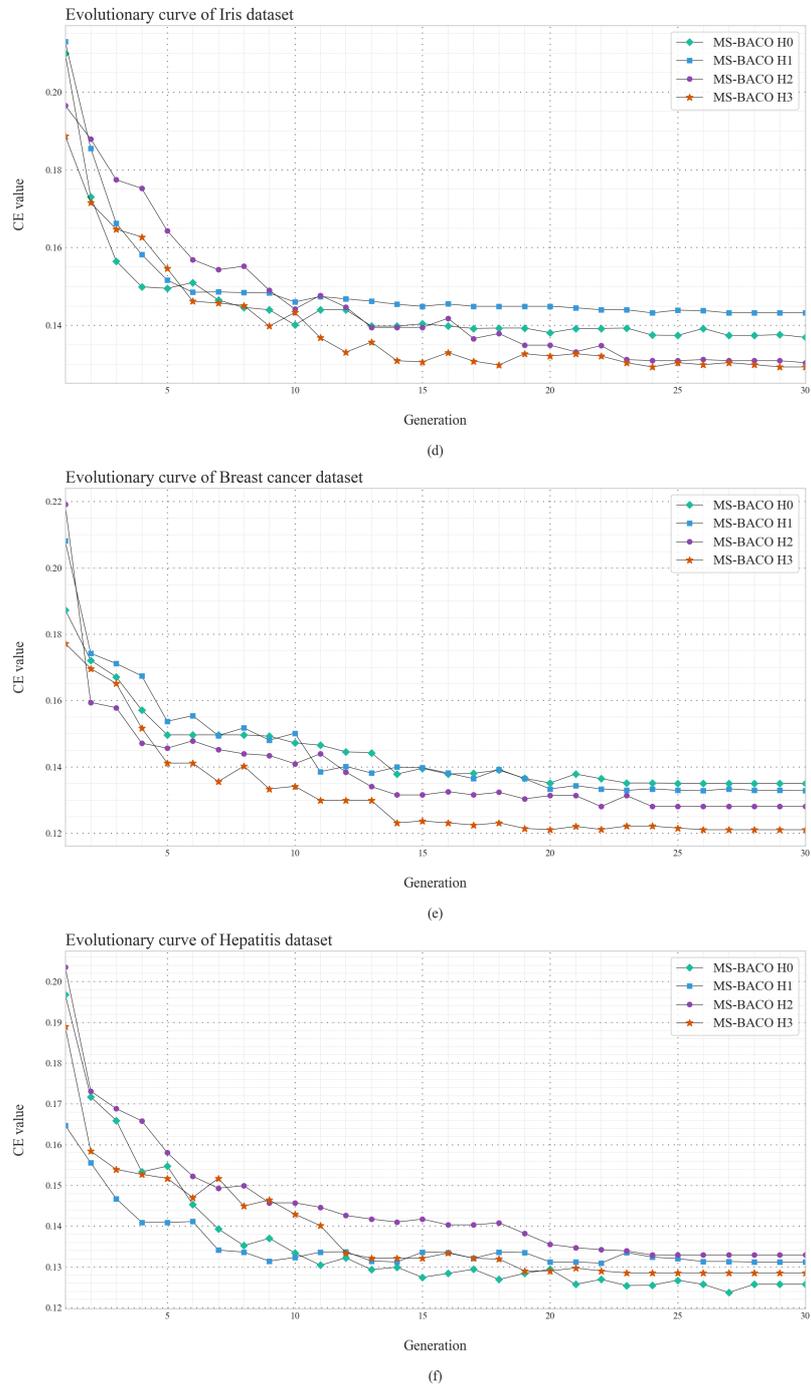

Fig. 9. Average evolutionary curve of datasets including, Iris (d), Breast cancer (e), Hepatitis (f).

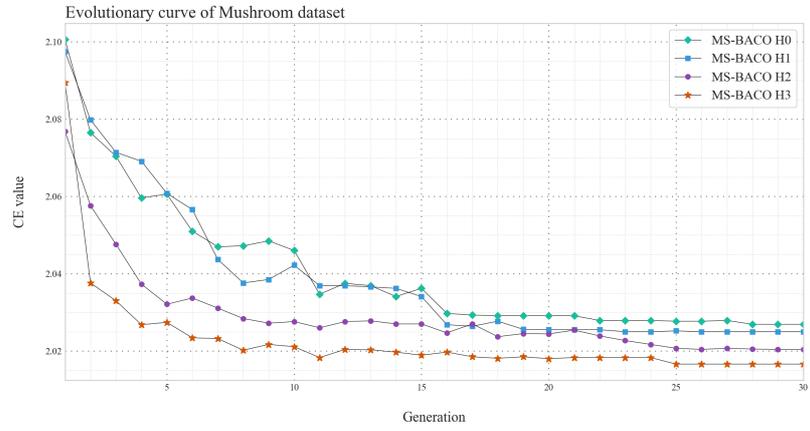

(g)

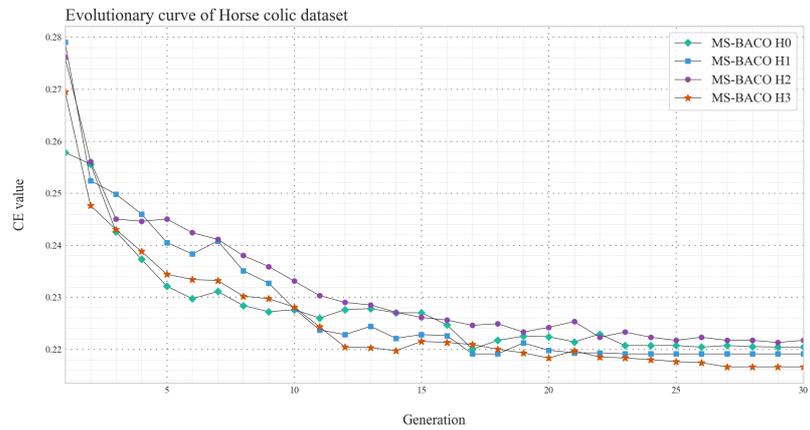

(h)

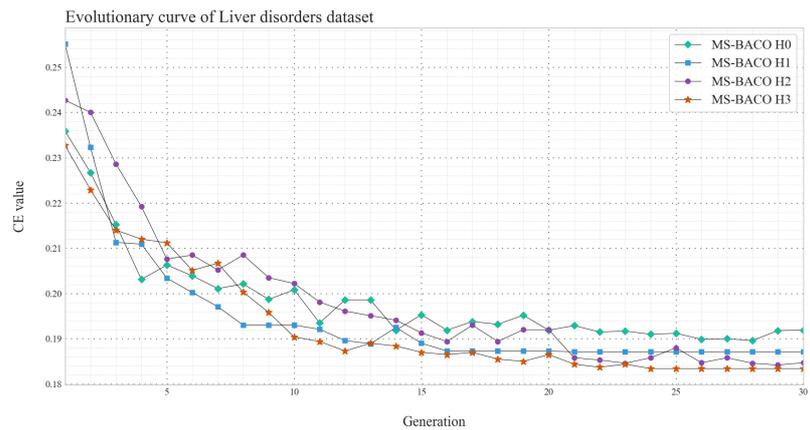

(i)

Fig. 10. Average evolutionary curve of datasets including, Mushroom (g), Horse colic (h), Liver disorders (i).

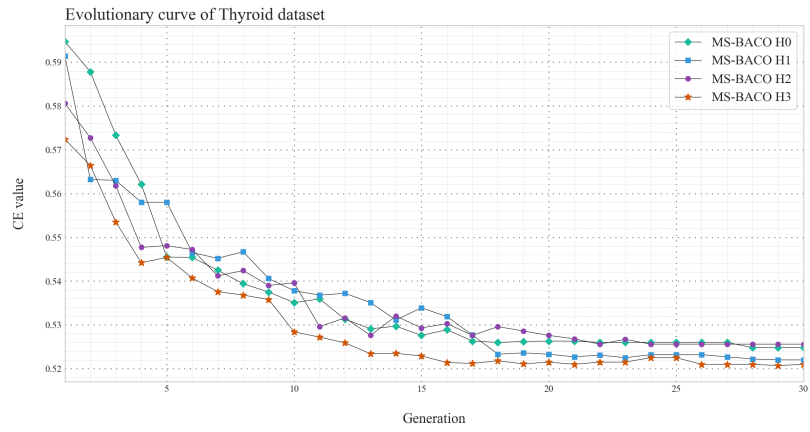

(j)

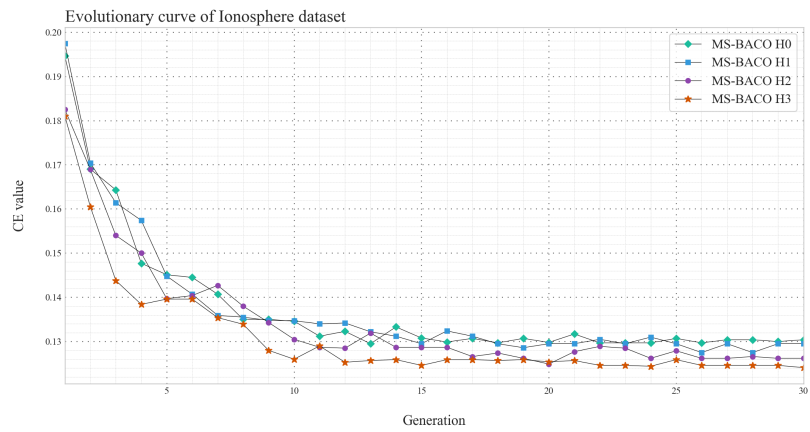

(k)

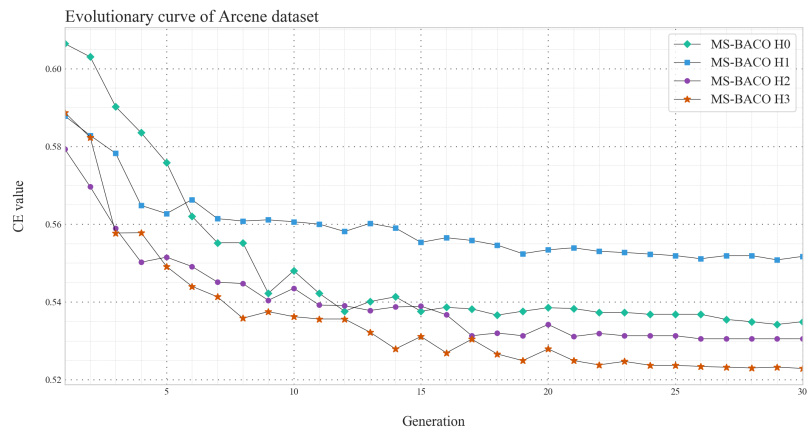

(l)

Fig. 11. Average evolutionary curve of datasets including, Thyroid (j), Ionosphere (k), Arcene (l).

### 4.3.  *Comparison with existing works*

To ensure optimal performance and efficiency of the optimization process in MS-BACO, performance of the algorithm is compared with some of the prominent node pruning, growing, hybrid and evolutionary algorithms from Fig. 2 such as, CPNN [15], HCPS [17], N2PS [20], IPSONet [26], HMOEN-L2 [29], Xing [21], that employ related techniques to MS-BACO including sensitivity measure, correlation analysis, state-space search, and also some of other structure optimization algorithms which are, SRS-N [37], C-Mantec [38], EL-ELM [32].

The empirical results are summarized in Table 4, where the average classification accuracy on test data and remaining number of hidden neurons for each algorithm and also, a Fixed FNN structure (F-FNN) with 50 hidden neurons are comparable. Based on analyzing data in Table 4, several comments can be made:

- In all of the problems that result is available to compare CPNN, HCPS, IPSONet, HMOEN-L2, Xing and EL-ELM algorithms to MS-BACO algorithm (i.e., Diabetes, Iris, Hepatitis, Breast cancer, Horse colic, Liver disorders, Yeast, Arcene), the proposed algorithm could obtain FNN model with lower size and higher averaged classification accuracy.

- Furthermore, results of SRS-N and MS-BACO algorithms are comparable for 7 problems including, Diabetes, Breast cancer, Iris, Wine, Mushroom, Liver disorders, Thyroid. MS-BACO algorithm outperforms SRS-N algorithm in 4 problems in terms of both model accuracy and compactness. Moreover, MS-BACO algorithm also achieves higher accuracy in Liver disorders and simpler model in Breast cancer problem.

- MS-BACO algorithm also reaches better classification accuracy in all the 4 comparable problems with N2PS algorithm (i.e., Breast cancer, Iris, Hepatitis and Ionosphere), and the final number of remaining hidden neurons in the network is lower than N2PS in Iris and Ionosphere problems.

- There are 5 comparable problems for ensuring about the better performance of the proposed algorithm compared to C-Mantec algorithm (i.e., Breast cancer, Mushroom, Horse colic, Thyroid and Ionosphere), MS-BACO algorithm has better accuracy in all of these problems except Mushroom problem while accuracies of both algorithms are the same. Additionally, MS-BACO algorithm also achieves more compact network in Horse colic and Thyroid problems.

Altogether, MS-BACO algorithm is verified to have a better performance in all four mutual analyses that took place on the joint problems.

### 4.4.  *Discussion on MS-BACO*

Previously, the results obtained by MS-BACO confirm that the proposed algorithm can outperform other algorithms. Here, we briefly explain some of the salient characteristics of MS-BACO algorithm that bring on this kind of result.

**Table 4.** Performance comparison of MS-BACO algorithm (using H3 heuristic design) with presented algorithms.

| Algorithm | Classification accuracy (%) | | Final No. of neurons | | Algorithm | Classification accuracy (%) | | Final No. of neurons | |
|---|---|---|---|---|---|---|---|---|---|
| | Mean | Std. | Mean | Std. | | Mean | Std. | Mean | Std. |
| **Diabetes** | | | | | **Yeast** | | | | |
| **MS-BACO** | **79.71** | 1.12 | **4.29** | 1.42 | **MS-BACO** | **59.14** | 1.31 | **11.23** | 1.67 |
| **F-FNN** | 79.23 | 1.74 | 50 | 0 | **F-FNN** | 58.76 | 1.64 | 50 | 0 |
| **HCPS** | 79.44 | 0.44 | 5 | - | **HCPS** | 58.80 | 2.75 | 18 | - |
| **HMOEN-L2** | 78.45 | 1.2 | 7.51 | 2.4 | **EL-ELM a** | 53.01 | 1.26 | 161 | 8 |
| **SRS-N** | 76.73 | 4.86 | 4.97 | 1.03 | **CPNN** | 52.81 | 1.56 | 42 | 5 |
| **IPSONet** | 76.68 | 2.0 | 4.9 | 1.2 | - | - | - | - | - |
| **Breast cancer** | | | | | **Iris** | | | | |
| **MS-BACO** | 97.63 | 0.82 | 2.36 | 0.79 | **MS-BACO** | **98.91** | 0.40 | **1.86** | 0.61 |
| **F-FNN** | 97.28 | 0.86 | 50 | 0 | **F-FNN** | 98.14 | 0.64 | 50 | 0 |
| **SRS-N** | **98.65** | 0.18 | 2.63 | 0.37 | **SRS-N** | 98.87 | 0.30 | 2.01 | 1.39 |
| **N2PS** | 97.10 | - | 2 | - | **Xing** | 98.67 | - | 2 | - |
| **IPSONet** | 97.07 | 0.50 | 4.7 | 1.00 | **N2PS** | 98.67 | - | 3 | - |
| **C-Mantec** | 96.86 | 1.19 | **1** | 0.00 | **HCPS** | 98.33 | 1.50 | 4 | - |
| **Xing** | 96.78 | - | 3 | - | **HMOEN-L2** | 98.00 | 1.84 | 3.08 | 0.86 |
| **Wine** | | | | | **Hepatitis** | | | | |
| **MS-BACO** | 98.94 | 1.04 | 5.29 | 1.20 | **MS-BACO** | **88.29** | 2.77 | 4.33 | 1.53 |
| **F-FNN** | 98.90 | 1.15 | 50 | 0 | **F-FNN** | 87.64 | 2.43 | 50 | 0 |
| **SRS-N** | **99.27** | 0.12 | **4.63** | 1.36 | **N2PS** | 86.40 | - | **3** | - |
| **Xing** | 98.89 | - | 6 | - | **Xing** | 84.62 | - | 8 | - |
| **HCPS** | 98.78 | 2.43 | 6 | - | **HMOEN-L2** | 80.30 | 4.8 | 11.38 | 3.26 |
| **Mushroom** | | | | | **Horse colic** | | | | |
| **MS-BACO** | **99.98** | 0.02 | 5.13 | 2.98 | **MS-BACO** | **98.54** | 1.32 | **7.00** | 2.55 |
| **F-FNN** | 99.73 | 0.55 | 50 | 0 | **F-FNN** | 97.70 | 2.01 | 50 | 0 |
| **C-Mantec** | **99.98** | 0.04 | **1.00** | 0.00 | **HMOEN-L2** | 98.38 | 1.31 | 7.20 | 5.05 |
| **SRS-N** | 99.91 | 0.09 | 7.50 | 2.30 | **C-Mantec** | 67.79 | 5.71 | 9.40 | 0.93 |
| **Liver disorders** | | | | | **Thyroid** | | | | |
| **MS-BACO** | **77.04** | 2.74 | 6.11 | 2.22 | **MS-BACO** | 97.71 | 0.85 | **4.93** | 1.24 |
| **F-FNN** | 76.40 | 3.09 | 50 | 0 | **F-FNN** | **97.73** | 0.30 | 50 | 0 |
| **SRS-N** | 76.23 | 4.18 | **4.33** | 1.37 | **SRS-N** | 97.65 | 1.32 | 6.30 | 0.70 |
| **HMOEN-L2** | 68.00 | 2.94 | 6.83 | 1.23 | **C-Mantec** | 94.16 | 0.51 | 5.00 | 0.00 |
| **Ionosphere** | | | | | **Arcene** | | | | |
| **MS-BACO** | **95.24** | 0.72 | 3.49 | 1.15 | **MS-BACO** | **90.03** | 1.44 | **16.75** | 4.12 |
| **F-FNN** | 95.11 | 1.00 | 50 | 0 | **F-FNN** | 89.80 | 1.24 | 50 | 0 |
| **N2PS** | 94.90 | - | 4 | - | **CPNN** | 87.41 | 0.97 | 132 | 12 |
| **C-Mantec** | 87.44 | 0.06 | **2.00** | 0.00 | **EL-ELM a** | 87.36 | 1.11 | 351 | 27 |

[a] indicates that the result is denoted in [15], all other results are from original papers.

- indicates no result is mentioned in original papers.

MS-BACO analyzes both contribution and correlation of hidden neurons for selecting the optimal model, unlike algorithms that utilize one of them alone. This leads to MS-BACO being capable of deselect more irrelevant neurons and achieving smaller network. Moreover, the algorithm outstrips other evolutionary computation algorithms that try to select network by searching the state-space without considering these measures. To demonstrate how the proposed algorithm modifies and revises the initial network, we contemplate the results derived from an experiment on Yeast dataset. The algorithm starts with the initial network with 50 hidden neurons and at the end, a network with 11 hidden neurons is suggested by the algorithm. Histogram and kernel density estimation (KDE)
of the correlation coefficient absolute values between hidden neurons in initial and final
networks are displayed in Fig. 12. As this figure depicts, final network has a lower number
of neurons with much less similarity among them, more similar to the network with 4
hidden neurons in Fig. 1.

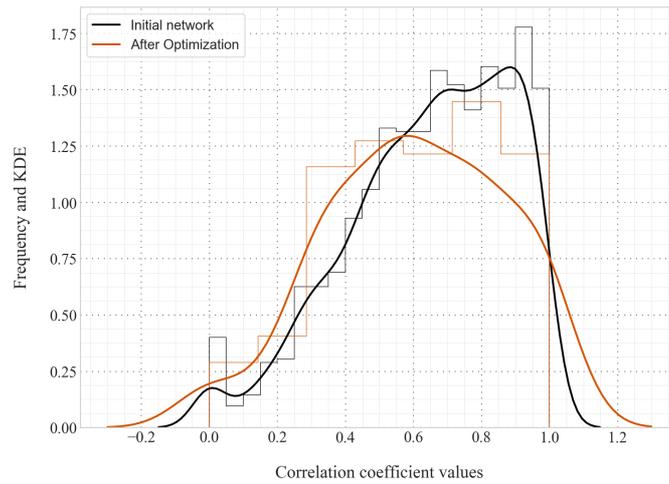

Fig. 12. Comparison of histogram and KDE of correlation coefficient absolute values between hidden neurons
in initial and final network.

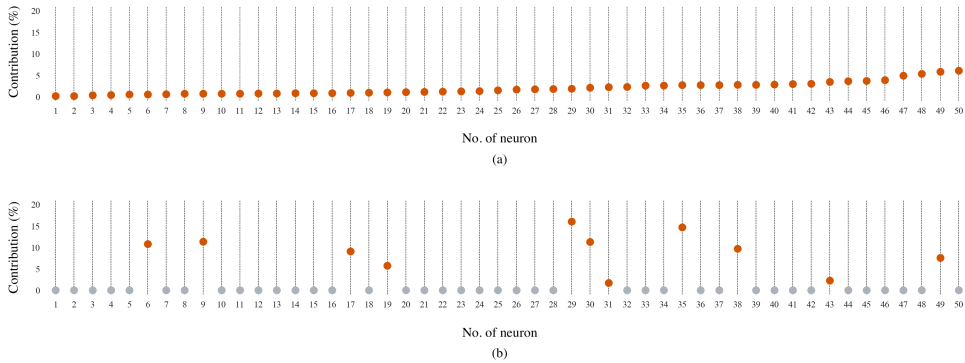

Fig. 13. Comparison of contribution percentage of hidden neurons in initial (a) and final (b) networks.

Meanwhile, Fig. 13.a shows the contribution percentage of hidden neurons in initial
network and Fig. 13.b shows the selected neurons also as well as their contribution
percentage in the optimized network. According to this figure, although most of the
selected neurons have relatively higher contribution in initial network. But unlike the
previous proposed algorithms that use sensitivity measure technique to eliminate nodes with lower saliency than a specified threshold, MS-BACO does not necessarily remove all of the neurons with lowest contribution percentage. Meanwhile, defining the CE as the objective function helps the model selection to take place with a kind of look-ahead strategy, that tries to select the best combination of neurons that reduce the learner error. In fact, sometimes the less contributed neurons have a better set of weight connections that is well worth to be remained in the network.

## 5. Conclusion

We presented a novel algorithm based on binary ant colony optimization for obtaining more accurate FNN model with lower number of hidden neurons which provides right level of neural complexity for a given problem. Optimizing the size of hidden layer also results in more efficiency in forward and backward propagations. By treating each neuron as a graph node, FNN model selection is contemplated as the problem of finding the shortest path, where the cost of the paths is cross-entropy error of the model, and agents try to respectively explore for the low-cost paths on the graph. In this case, unlike other algorithms, selection of neuron subsets and removal of irrelevant neurons with arbitrary numbers is assigned to algorithm itself instead of using pre-defined problem dependent thresholds. By calculating both correlation between hidden neurons and their contribution in overall network output variations using sensitivity analysis, the proposed algorithm mostly tries to select the most distinct hidden neurons with high contribution. Simulation results demonstrates that the proposed algorithm has better performance among other previous proposed algorithms.

## References

1       Narendra KS, Parthasarathy K. Identification and control of dynamical systems using neural networks. *IEEE Transactions on neural networks*. 1990; **1**: 4-27.

2       Achili B, Daachi B, Ali-Cherif A, Amirat Y. Combined multi-layer perceptron neural network and sliding mode technique for parallel robots control: an adaptive approach. *Neural Networks, 2009 IJCNN 2009 International Joint Conference on*: IEEE, 2009; 28-35.

3       Yu L, Wang S, Lai KK. A neural-network-based nonlinear metamodeling approach to financial time series forecasting. *Applied Soft Computing*. 2009; **9**: 563-74.

4       Karabacak K, Cetin N. Artificial neural networks for controlling wind–PV power systems: A review. *Renewable and Sustainable Energy Reviews*. 2014; **29**: 804-27.

5       Jan B, Farman H, Khan M*, et al.* Deep learning in big data Analytics: A comparative study. *Computers & Electrical Engineering*. 2017.

6       Bishop CM. Regularization and complexity control in feed-forward networks. 1995.

7       Kon MA, Plaskota L. Information complexity of neural networks. *Neural Networks*. 2000; **13**: 365-75.

8       Goodfellow I, Bengio Y, Courville A, Bengio Y. *Deep learning*: MIT press Cambridge, 2016.

9       Huang G-B, Zhu Q-Y, Siew CK. Real-time learning capability of neural networks. *IEEE Trans Neural Networks*. 2006; **17**: 863-78.

10      Rumelhart DE, Hinton GE, Williams RJ. Learning representations by back-propagating errors. *nature*. 1986; **323**: 533.

11      Zeng X, Yeung DS. Hidden neuron pruning of multilayer perceptrons using a quantified sensitivity measure. *Neurocomputing*. 2006; **69**: 825-37.

12      Dam M, Saraf DN. Design of neural networks using genetic algorithm for on-line property estimation of crude fractionator products. *Computers & chemical engineering*. 2006; **30**: 722-29.

13      Lin J-H, Vitter JS. Complexity results on learning by neural nets. *Machine Learning*. 1991; **6**: 211-30.

14      Lauret P, Fock E, Mara TA. A node pruning algorithm based on a Fourier amplitude sensitivity test method. *IEEE transactions on neural networks*. 2006; **17**: 273-93.

15      Han H-G, Qiao J-F. A structure optimisation algorithm for feedforward neural network construction. *Neurocomputing*. 2013; **99**: 347-57.

16      Thomas P, Suhner M-C. A new multilayer perceptron pruning algorithm for classification and regression applications. *Neural Processing Letters*. 2015; **42**: 437-58.

17      Qiao J, Li S, Han H, Wang D. An improved algorithm for building self-organizing feedforward neural networks. *Neurocomputing*. 2017; **262**: 28-40.

18      Engelbrecht AP. A new pruning heuristic based on variance analysis of sensitivity information. *IEEE transactions on Neural Networks*. 2001; **12**: 1386-99.

19      Xu J, Ho DW. A node pruning algorithm based on optimal brain surgeon for feedforward neural networks. *International Symposium on Neural Networks*: Springer, 2006; 524-29.

20      Augasta MG, Kathirvalavakumar T. A novel pruning algorithm for optimizing feedforward neural network of classification problems. *Neural processing letters*. 2011; **34**: 241.

21      Xing H-J, Hu B-G. Two-phase construction of multilayer perceptrons using information theory. *IEEE Transactions on Neural Networks*. 2009; **20**: 715-21.

22      Zhang Z, Chen Q, Qiao J. A merging and splitting algorithm based on mutual information for design neural networks. *Bio-Inspired Computing: Theories and Applications (BIC-TA), 2010 IEEE Fifth International Conference on*: IEEE, 2010; 1268-72.

23      Islam MM, Sattar MA, Amin MF, Yao X, Murase K. A new adaptive merging and growing algorithm for designing artificial neural networks. *IEEE Transactions on Systems, Man, and Cybernetics, Part B (Cybernetics)*. 2009; **39**: 705-22.

24      Abbas HM. A decorrelation approach for pruning of multilayer perceptron networks. *IAPR Workshop on Artificial Neural Networks in Pattern Recognition*: Springer, 2014; 12-22.

25      Oh S-H. Hidden node pruning of multilayer perceptrons based on redundancy reduction. *International Conference on Hybrid Information Technology*: Springer, 2011; 245-49.

26      Yu J, Xi L, Wang S. An improved particle swarm optimization for evolving feedforward artificial neural networks. *Neural Processing Letters*. 2007; **26**: 217-31.

27      Qi C, Fourie A, Chen Q. Neural network and particle swarm optimization for predicting the unconfined compressive strength of cemented paste backfill. *Construction and Building Materials*. 2018; **159**: 473-78.

28      Benardos P, Vosniakos G-C. Optimizing feedforward artificial neural network architecture. *Engineering Applications of Artificial Intelligence*. 2007; **20**: 365-82.

29      Goh C-K, Teoh E-J, Tan KC. Hybrid multiobjective evolutionary design for artificial neural networks. *IEEE Transactions on Neural Networks*. 2008; **19**: 1531-48.

30      Garro BA, Rodríguez K, Vazquez RA. Designing artificial neural networks using differential evolution for classifying DNA microarrays. *Evolutionary Computation (CEC), 2017 IEEE Congress on*: IEEE, 2017; 2767-74.

31      Wang N, Han M, Dong N, Er MJ. Constructive multi-output extreme learning machine with application to large tanker motion dynamics identification. *Neurocomputing*. 2014; **128**: 59-72.

32      Huang G-B, Chen L. Enhanced random search based incremental extreme learning machine. *Neurocomputing*. 2008; **71**: 3460-68.

33      Zhang R, Lan Y, Huang G-b, Xu Z-B. Universal approximation of extreme learning machine with adaptive growth of hidden nodes. *IEEE Transactions on Neural Networks and Learning Systems*. 2012; **23**: 365-71.

34      Miche Y, Sorjamaa A, Bas P, Simula O, Jutten C, Lendasse A. OP-ELM: optimally pruned extreme learning machine. *IEEE transactions on neural networks*. 2010; **21**: 158-62.

35      Qiao J, Li F, Han H, Li W. Constructive algorithm for fully connected cascade feedforward neural networks. *Neurocomputing*. 2016; **182**: 154-64.

36      Li F, Qiao J, Han H, Yang C. A self-organizing cascade neural network with random weights for nonlinear system modeling. *Applied soft computing*. 2016; **42**: 184-93.

37      Yang J, Ma J. A structure optimization framework for feed-forward neural networks using sparse representation. *Knowledge-Based Systems*. 2016; **109**: 61-70.

38      Subirats JL, Franco L, Jerez JM. C-Mantec: A novel constructive neural network algorithm incorporating competition between neurons. *Neural Networks*. 2012; **26**: 130-40.

39      Kwok T-Y, Yeung D-Y. Constructive algorithms for structure learning in feedforward neural networks for regression problems. *IEEE transactions on neural networks*. 1997; **8**: 630-45.

40      Reed R. Pruning algorithms-a survey. *IEEE transactions on Neural Networks*. 1993; **4**: 740-47.

41      Hansen LK, Rasmussen CE. Pruning from adaptive regularization. *Neural Computation*. 1994; **6**: 1223-32.

42      Larsen J, Hansen LK, Svarer C, Ohlsson M. Design and regularization of neural networks: the optimal use of a validation set. *Neural Networks for Signal Processing [1996] VI Proceedings of the 1996 IEEE Signal Processing Society Workshop*: IEEE, 1996; 62-71.

43      McCulloch WS, Pitts W. A logical calculus of the ideas immanent in nervous activity. *The bulletin of mathematical biophysics*. 1943; **5**: 115-33.

44      Wenger E, Brozzoli C, Lindenberger U, Lövdén M. Expansion and Renormalization of Human Brain Structure During Skill Acquisition. *Trends in cognitive sciences*. 2017; **21**: 930-39.

45      Fu M, Zuo Y. Experience-dependent structural plasticity in the cortex. *Trends in neurosciences*. 2011; **34**: 177-87.

46      Kilgard MP. Harnessing plasticity to understand learning and treat disease. *Trends in neurosciences*. 2012; **35**: 715-22.

47      Quallo M, Price C, Ueno K, *et al.* Gray and white matter changes associated with tool-use learning in macaque monkeys. *Proceedings of the National Academy of Sciences*. 2009; **106**: 18379-84.

48      Hiroyasu T, Miki M, Ono Y, Minami Y. Ant colony for continuous functions. *The Science and Engineering, Doshisha University*. 2000; **20**.

49      Dorigo M, Di Caro G. Ant colony optimization: a new meta-heuristic. *Evolutionary Computation, 1999 CEC 99 Proceedings of the 1999 Congress on*: IEEE, 1999; 1470-77.

50      Hornik K, Stinchcombe M, White H. Multilayer feedforward networks are universal approximators. *Neural networks*. 1989; **2**: 359-66.

51      Cybenko G. Approximation by superpositions of a sigmoidal function. *Mathematics of control, signals and systems*. 1989; **2**: 303-14.

52      Zhou P, Austin J. Learning criteria for training neural network classifiers. *Neural computing & applications*. 1998; **7**: 334-42.

53      Golik P, Doetsch P, Ney H. Cross-entropy vs. squared error training: a theoretical and experimental comparison. *Interspeech*2013; 1756-60.

54      Saltelli A, Sobol IM. About the use of rank transformation in sensitivity analysis of model output. *Reliability Engineering & System Safety*. 1995; **50**: 225-39.

55      Saltelli A, Annoni P, Azzini I, Campolongo F, Ratto M, Tarantola S. Variance based sensitivity analysis of model output. Design and estimator for the total sensitivity index. *Computer Physics Communications*. 2010; **181**: 259-70.

56      Saltelli A, Tarantola S, Campolongo F, Ratto M. *Sensitivity analysis in practice: a guide to assessing scientific models*: John Wiley & Sons, 2004.

57      Jensen R, Shen Q. Finding rough set reducts with ant colony optimization. *Proceedings of the 2003 UK workshop on computational intelligence*2003; 15-22.

58      Dorigo M, Stützle T. *Ant Colony Optimization*: BRADFORD BOOK, 2004.

59      Pérez F, Granger BE. IPython: a system for interactive scientific computing. *Computing in Science & Engineering*. 2007; **9**.

60      Oliphant TE. Python for scientific computing. *Computing in Science & Engineering*. 2007; **9**.

61      Asuncion A, Newman D. UCI machine learning repository. 2007.

62        Rosenblatt M. Remarks on some nonparametric estimates of a density function. *The Annals of Mathematical Statistics*. 1956: 832-37.